# Genre determining prediction: Non-standard TAM marking in football language

**Jakob Egetenmeyer**

Affiliation: SFB 1252 / Department of Romance Studies, University of Cologne, Cologne, Germany

Email: j.egetenmeyer@uni-koeln.de

**Abstract**

German and French football language display tense-aspect-mood (TAM) forms which differ from the TAM use in other genres. In German football talk, the present indicative may replace the pluperfect subjunctive. In French reports of football matches, the imperfective past may occur instead of a perfective past tense-aspect form. We argue that the two phenomena share a functional core and are licensed in the same way, which is a direct result of the genre they occur in. More precisely, football match reports adhere to a precise script and specific events are temporally determined in terms of objective time. This allows speakers to exploit a secondary function of TAM forms, namely, they shift the temporal perspective. We argue that it is on the grounds of the genre that comprehenders predict the deviating forms and are also able to decode them.

We present various corpus studies where we explore the functioning of these phenomena in order to gain insights into their distribution, grammaticalization and their functioning in discourse. Relevant factors are Aktionsart properties, rhetorical relations and their interaction with other TAM forms. This allows us to discuss coping mechanisms on the part of the comprehender. We broaden our understanding of the phenomena, which have only been partly covered for French and up to now seem to have been ignored in German.

# 1. Introduction

In German and French football language we find tense-aspect-mood (TAM) forms which deviate from the TAM use in similar structural environments in other genres. We set out to investigate these phenomena in depth using various online corpora. Both phenomena are verbal in nature. They also share the property that the form used lacks a relevant feature which, by contrast, a competing form would express. However, they differ with respect to the relevant referential domain and therefore also in terms of the paradigmatic position of the forms in the language system. On the one hand, in German football language, the present indicative may take the place of the pluperfect subjunctive, which impacts the world reference coordinate. On the other hand, French football reports contain uses of the imperfective past tense-aspect form (*imparfait*) expressing sequences of events which would be realized by a perfective past in other genres. The deviation is thus located on the level of temporal reference. Although we might suspect erroneous interpretations in terms of truth values (for German) and temporal sequentiality (for French), the deviating forms do not seem to pose difficulties for comprehenders. Interestingly, the phenomena share important parallels in the licensing and functioning of the deviations.

The analysis of the two languages indicates that it is specifically the genre which enables speakers to interpret the conflicting forms correctly. We argue that due to the properties of the genre, the precise semantic and script-based expectations outweigh the effect of the deviating TAM forms. Football reports are special in several respects. There is a reduced set of typical events, they adhere to a well-entrenched script, and the events referred to may also be temporally determined in terms of an objective time line. We argue that these properties allow the speakers to exploit secondary functions of TAM forms. They access the temporal coordinates and shift the temporal perspective (see de Saussure and Sthioul, 1999). Thus, the hypothesis is that in the case of football language, deviating TAM forms are predicted on the grounds of genre.

There has been some work on tense in English football language (see Walker, 2008), and also on the imperfectives occurring in French football language (see Labeau, 2004; Labeau, 2007; Egetenmeyer, in press). By contrast, the German phenomenon does not seem to have been covered in the literature. Furthermore, as we show in the present contribution, the French imperfective may in fact appear as the only inflected verb form in entire newspaper articles on football. Our investigation deepens the understanding of the different phenomena in German and French. On the grounds of several corpus studies with different kinds of data, we carve out the relevant properties and determine how the phenomena function within discourse. In order to achieve this aim, we address five research questions. The first two focus on the quality of the phenomena in the linguistic system. (i) In what kind of texts and contexts do we find the deviating TAM forms and what can we say about their frequency? (ii) What is the status of the forms within the given linguistic system? The other three questions relate to the speaker's intentions and the role of the genre for the decoding of the message on the part of the comprehender. (iii) For what reason do speakers apply deviant TAM forms? (iv) How do comprehenders cope with the deviating TAM forms and how do they resolve the missing information? (v) Finally, what do these phenomena tell us about predictive language processing?

We proceed as follows. In Section 2, we introduce the theoretical background. We show in what way the forms in question deviate from the standard TAM use (Sections 2.1 and 2.2). We discuss the commonalities of the uses (Section 2.3) and the role of predictions in their decoding (Section 2.4). In Section 3, we describe the corpora used in this study and how we approach them. First, we focus on the German data (Section 3.1), then on the French (Section 3.2). The results are presented in Section 4. We begin with what we found on the Aktionsart of the verbs involved (Section 4.1) and continue with properties pertaining to the discourse level (Section 4.2). In Section 5, we discuss our results and summarize what we have found regarding the research questions mentioned above. Furthermore, we draw up observations on possible further research.

## 2. Theoretical background

In this section, we present the relevant theoretical background necessary to discuss the data of interest properly. This is important for the following reasons. We account for linguistic phenomena in two languages, German and French. The two languages pertain to different language families and the functioning of TAM categories is not identical. We focus on two different phenomena, mood choice in German and tense-aspect choice in French. However, we approach both of them with the same interest, namely, the role of genre for their decoding. As comprehenders understand them without difficulties, we argue that they actually predict such forms to occur in football language. The object of study, especially the German TAM forms of interest, and the way we address the data is to some extent new; for instance, TAM research does not normally include the factor of genre, and therefore needs theoretical backing. Thus, finally, apart from the linguistic phenomena and insights into their commonalities, we also introduce predictive language processing in this section.

The subsections are ordered in the following way. In Section 2.1, we present the German phenomenon we investigate. In Section 2.2, we continue with the French counterpart. With regard to both languages, we specify the linguistic variety in which we find the phenomena in question. In the first two subsections, we will mention important characteristics. Section 2.3 discusses the common core of the functioning of the German and French phenomena. It shows that they are both licensed by the specific properties of the genre. Thus, the section also motivates why we account for the two very different phenomena in a parallel fashion. Section 2.4 replenishes the theoretical background with insights into the factor of prediction.

## 2.1 German TAM forms: Present indicative substituting pluperfect subjunctive

The formal deviation we are interested in with respect to German occurs in conditional clauses. More precisely, we are interested in structures of past reference with a counterfactual or irrealis reading but which are realized with a present indicative verb form. Thus, this first puzzle forms part of the realm of modality. According to the basic definition of Palmer (2001, p. 1), "[m]odality is concerned with the status of the proposition that describes the event". Although the definition has been said to be imprecise (see Salkie, 2009, p. 79), it may serve to emphasize a crucial point, as it does not specify how this status is brought about. In German conditional clauses, the protasis may or may not be introduced by the conjunction *wenn* ('if'). The alternatives *falls* ('if') and *sofern* ('provided that') are not compatible with the counterfactual reading (see Zifonun et al., 1997, p. 2280) and might be used as a test battery. When there is no conjunction, the subordinate clause is realized as a verb-first clause (see Zifonun et al., 1997, p. 2281). The apodosis may but does not have to involve the adverbial *dann* ('then'). Although, in general, certain cases of syncretism between indicative and conjunctive exist (see Zifonun et al., 1997, p. 1739-1743), the verbal forms occurring in the protasis and the apodosis indicate the propositional status in a largely unequivocal fashion (see Zifonun et al., 1997, p. 1745-1746). This is a more general phenomenon, which is not restricted to German; for instance, Portner (2009, p. 221-247) discusses the interplay between modality and tense-aspect forms for English. As Zifonun et al. (1997, p. 1745) put it, when there is an indicative in the protasis this may yield a hypothetical reading, but counterfactuality is generally ruled out. By contrast, counterfactuality is normally expressed by means of a pluperfect subjunctive in the subordinate clause and a past subjunctive (also called subjunctive II; see, for instance, Fabricius-Hansen, 1999 for the German verb paradigm) or another pluperfect subjunctive in the main clause (Zifonun et al., 1997, p. 1745-1746). Declerck (2011, p. 28) calls this phenomenon "'[m]odal backshifting' (or 'formal distancing')". Importantly, its functioning differs from the backshifting of verb forms found in indirect speech, and therefore should not be confused with it (see Declerck, 2011, p. 28). Zifonun et al. (1997, p. 1746) present the following example.

(1) *Wenn die Sängerin gelächelt hätte, {wären wir glücklich / wären wir glücklich gewesen}.* (Zifonun et al., 1997, p. 1746, adapted)

'If the singer had smiled, we {would be / would have been} happy.'

If we focus on the apodosis in (1), the first variant yields a co-temporal reference with regard to the moment of speech, while the preferred reading of the second possibility is one of past reference (see Leirbukt, 2008, discussed below). Still, Zifonun et al. (1997, p. 1747-1748) mention the less typical possibility that a counterfactual reading may arise with mixed forms in which one half of the structure contains a past subjunctive form while the other shows an indicative. However, as becomes apparent in the examples cited in Zifonun et al. (1997, p. 1747), a strong contextual determination is necessary and the structure appears to be marked.

In his introductory section, Leirbukt (2008, p. 1-6) presents the various possible ways of expressing potentiality and counterfactuality in German. As he shows, the distinction is not independent of the temporal localization (see Leirbukt, 2008, p. 1-6). Lewis (1979) takes a philosophical stance and discusses the role the divergence of an invariant past, as opposed to an undetermined and therefore flexible future, has on counterfactuals. Now, as already noted, when the German pluperfect subjunctive occurs in a past context, a counterfactual reading is typically realized; however, it is not the only possibility, as Leirbukt (2008, p. 27 with further references) shows, although he focusses on non-past temporal reference (see Leirbukt, 2008, p. 6).

However, we assume that in general, the argument cannot be inverted. So, in order to express a counterfactual reading in the past, the pluperfect subjunctive should be necessary. But football language teaches us otherwise, as example (2) shows. Up to this point, in our inquiry into the research literature, we have neither found reference to the present indicative substituting the pluperfect subjunctive in general, nor to its highly interesting use in football language. Furthermore, we have not come across this phenomenon in French, for which we analyze another phenomenon, as presented in the following section (see Becker, 2014 for a description of mood in Romance languages).

(2) *Latza [...] dachte nach dem 1:1 in der Domstadt an die vergebene Großchance des eingewechselten Teamkollegen Robin Quaison in der letzten der 97 Minuten: „Wenn er den* <u>*macht,*</u> <u>*heulen*</u> *hier 50.000 rum – und wir* <u>*freuen*</u> *uns. Schade.“* (FR 1)

'After the 1:1 in Cologne, Latza thought about the missed big chance of the substitute teammate Robin Quaison in the last of the 97 minutes: "If he had converted (lit.: converts) that one, 50,000 people would have cried (lit.: cry) with disappointment – and we would have been (lit.: are) happy. Too bad."'

As the example shows, a clearly counterfactual proposition is conveyed by means of verbs marked by the present indicative in both the protasis and the apodosis of the conditional clause. To our knowledge, beyond football language this is ruled out. Crucially, the use shows a tendency towards being an oral phenomenon. However, as shown by the above example and example (3) below, it is brought into the written form within direct quotes. In example (2), the occurrence shows two further markers of genre and sociolinguistic status, namely, the structure in the protasis (*den machen*, 'convert that one') is a typical expression in the language of football, with a noticeable marker for the language of proximity in the terms of Koch and Oesterreicher (2011). Furthermore, in the apodosis we find the verb *rumheulen* ('whine') which is marked as colloquial. According to our data, example (2) may be seen as a quite typical instance of the phenomenon. Among the factors to discuss are the following. The protasis features a telic event expression (*den machen*, 'convert that one'). The apodosis expresses an activity (*rumheulen*, 'whine') (and an additional state, *freuen*, 'are happy'). They show a rhetorical relation of consequence (see Asher and Lascarides, 2003, p. 169), which may be considered less typical.

However, apart from the genre restriction, the phenomenon is versatile. Most importantly, it has two different syntactic instantiations. The second type is exemplified in example (3), where the protasis lacking the conjunction *wenn* ('if') is realized as a verb-first sentence (see Zifonun et al., 1997, p. 2281).

(3) *[D]er Innenverteidiger, der sich noch immer über seine vergebene Kopfballchance im letzten WM-Gruppenspiel gegen Südkorea ärgert[, sagte]: „*<u>*Mache*</u> *ich das Tor,* *kommen wir gegen Südkorea weiter, dann wären viele Dinge sicherlich anders gelaufen.*" (Spiegel 1)

'The central defender, who is still upset about his missed header chance in the final World Cup group match against South Korea, said: "If I had scored (lit.: score), we would have succeeded (lit.: succeed) against South Korea to the next round, then many things would probably have gone differently."'

## 2.2 French TAM forms: Imperfective substituting perfective past

In classic literary French, two main past tense forms are used which express the aspectual distinction between perfective (*passé simple*) and imperfective (*imparfait*). Diachronically speaking, the *passé simple* has been extensively substituted in oral discourse and, to a certain extent, also in written discourse by the compound past (*passé composé*) (see Verkuyl et al., 2004, p. 253, 265-266). However, the opposition with the imperfective past is maintained (see Molendijk et al., 2004, p. 298; refinements can be found, however, in Facques, 2002). Thus, typically, series of past events are expressed by the simple past or the compound past, while the imperfective past tense-aspect form is used for co-occurring or background eventualities (see Kamp and Rohrer, 1983; Weinrich, 1964). However, in certain contexts, the imperfective past may substitute its perfective counterpart. This phenomenon is often called the *imparfait narratif* ('narrative imperfect'; see Bres, 2005; Gosselin, 1999; and others). Importantly, such occurrences are quite restricted in terms of syntactic, contextual and genre-related terms (see Caudal, submitted; Egetenmeyer, in press). Apart from literary texts and newspaper articles on politics (see Egetenmeyer, in press), we also find football reports among the genres which feature such imperfective tense-aspect uses (see Labeau, 2007, p. 220 who quotes Herzog, 1981, p. 67 as having noted the distribution quite early). Importantly, as Egetenmeyer (in press) underlines, its use in football reports is peculiar: It is the only genre which seems to allow for a full substitution of non-imperfective tense-aspect forms with the imperfective (however, see Facques, 2002, p. 115 for an example coming from a non-football related newspaper article which also shows a rather strong tendency to avoid non-imperfective tenses-aspect forms). More specifically, whole reports of football matches may be written using the *imparfait* where otherwise perfective (or non-imperfective) tense-aspect forms would be used. Interestingly, when comparing the situation with Spanish, which shows many parallels in the use of the corresponding imperfective past (see Escandell-Vidal, submitted with further references), we find that the usage in question is not paralleled there. For instance, as Quintero Ramírez and Carvajal Carvajal (2017, p. 229) indicate, in Mexican Spanish newspaper reports the imperfect past tense is not used to express sequences of eventualities.

Example (4), taken from Egetenmeyer (in press), is the beginning of a newspaper article reporting a football match. All five finite verb forms would be expected to be realized as compound past forms due to the expression of sequences of events. However, they are all marked by the imperfective, as are most of the other finite verbs in the rest of the article (see Egetenmeyer, in press).

(4) [1] *Le Blésois Gonçalves **était** le premier à se mettre en action (10^e^),* [2] *mais sa frappe **passait** juste à côté.* [3] *Les locaux **répondaient** de suite, avec une bonne tête de Maelbrancke,* [4] *mais le défenseur Radet **sauvait** sur sa ligne.* [5] *La réponse blésoise ne se **faisait** pas attendre [...].* (Sketch Engine: La Nouvelle République, 22.08.2016)

'[1] The Blesoisian Gonçalves was the first one to get into gear (10[th]), [2] but his shot just missed. [3] The locals responded immediately with a good header by Maelbrancke,

[4] but the defender Radet saved on the line. [5] The Blesoisian answer was not long in coming.'

As we will see, in such structures, we find a high proportion of verbs lexically expressing boundedness (see also Bres, 1999, p. 5), which we assume facilitates processing. As Egetenmeyer (in press) underlines, the specialty of this usage is that no temporal determination is necessary in order to license it. This contrasts with the use, for instance, of the typical narrative imperfect, which tends to co-occur with a temporal sentence adverbial under which it is embedded (see Egetenmeyer, in press) or similar uses appearing in relative clauses (see Caudal, submitted, who analyzes the examples in Bres, 2005). In those types of uses, we find a direct or indirect temporal determination of the expressed eventuality. It should be mentioned, however, that the above example does in fact contain an explicit (relative) temporal indication, namely, "10ᵉ [minute]" ('10th minute'). As our data show, such an indication is not necessary for the usage. However, as we will see in the following subsection, it makes a principle explicit which we assume to be relevant for the occurrence of imperfective tense-aspect forms in football reports, namely that it shows that a football match is measured in terms of an objective time. In the following section, we will go into more detail regarding the properties of the genre of football reports.

## 2.3 Commonalities: Football frame and shifted perspective time

In the two preceding subsections, we introduced the basic characteristics of the phenomena of interest. While they show certain parallels, they are also different in important respects. They share the basic property of pertaining to the verbal domain. In both languages, the TAM marking deviates from what would be expected in a different genre. Simply put, the marking would not withstand a normative stance. By contrast, an important difference is that the German TAM marking deviates in the realm of world reference, while the French counterpart shows its deviation with regard to temporal reference. Finally, they share two decisive properties, which also motivate their joint treatment. First, an important licensing factor for their realization, which is directly connected with the factor of genre, is given by the frame or the script of football matches. Second, the functioning of both phenomena can be explained as a shift in perspective time. In the following, we go into the details of these last two ideas.

A football match is conventionalized and functions according to a specific set of rules, of which at least the basic ones are known to the general public in the speech communities relevant for this paper (see also the interesting properties ascribed to football reports in newspapers by Engel and Labeau, 2005, p. 204-205, with reference to Grevisse, 1997, some of which, however, would need sociological verification). Therefore, (at least) the central information block of football language may be taken to show relevant features covered by accounts of scripts (see Schank and Abelson, 1977, p. 36-68) and frames (Fillmore, 1977; Fillmore, 2006). There are three further related properties of football matches which distinguish them, for instance, from the often-cited restaurant script. First, the non-generalized events of the football match are directly related to an objective time line. In a related matter, Engel and Labeau (2005, p. 215, with further references) remark that in football reports the events are often presented chronologically. Second, the matches of interest to a large group of people are televised. The large group of passive participants (i.e., viewers) and the factor of television broadcast further objectivizes the match, as there are many witnesses and the match can be (and is, in fact) recorded to be watched again. Importantly, the objective temporal determination of specific events is not circumstantial, but plays a decisive role for the match. For instance, it may have consequences for tactical planning. Furthermore, due to its role within the match, it has a high informative value in football reports and many other situations where football language is used. These very specific temporal properties have an influence on what needs to be conveyed in a relevant speech situation. We assume that it is due to such properties that the rigidity of certain components of the linguistic system may be attenuated. As a consequence, other functions may be exploited. As we will see, both languages make use of this principle in a similar way, although they diverge in terms of what exactly is modified.

In terms of discourse structural functioning (see Becker and Egetenmeyer, 2018 for our conception of temporal discourse structure), the two phenomena adhere to the same principle, namely, they show a shift in temporal perspective. For the French narrative imperfect, this has been discussed in a similar vein by Berthonneau and Kleiber (1999), de Saussure and Sthioul (1999) and Schrott (2012). Labeau (2006) mentions perspective specifically in the context of television talk, a category to which the genre of football reports partly pertains. According to Becker and Egetenmeyer (2018, p. 37, with reference to Guéron, 2015, p. 278), the perspective time "is the point in time where the text-internal origo is situated". While, in a standard narration of successive events expressed by means of verbs marked with the perfective past, the perspective time is anchored to the speech time, the eventive use of the *imparfait* in football reports is accompanied and licensed by a shift in perspective to the past. In terms of de Saussure and Sthioul (1999, p. 6), the perspective time is included in the run time of the event, which, however, may be determined more precisely as the location time corresponding to the event (see Becker and Egetenmeyer, 2018). With regard to football language, we have to keep in mind a further component which seems to be missing in the above-mentioned publications, namely, that this perspective time needs to be continuously updated as the events are narrated one after the other by means of verbs marked by the *imparfait*. A similar principle is described by de Swart (2007, p. 2282) with regard to the special use of the French present perfect in Camus' *L'Étranger*, which, according to her, is mirrored by the adverbials used in the text. Schrott (2012) emphasizes the underlying perception of the perspectivizing entity with the narrative *imparfait*. Envisioned in this way, this use of the *imparfait* could be understood as a means to bring the report closer to the speaker / hearer and thereby to render it livelier (see, however, Labeau, 2007, p. 220, who notes that the literary narrative *imparfait* with a temporal adverbial tends to render a passage rather clumsy). The actualization of the secondary (competing) perspectivizing function in the footballer's context might also be applied to the English narrative present perfect, described in the football context by Walker (2008), and even to the use of the present perfect in Australian police reports (see Ritz, 2010). These two publications, however, do not mention this interpretation.

Although the German phenomenon we are interested in does not pertain to the realm of tense-aspect but to the modal domain, it may also be interpreted in this way. In this interpretation, the speaker shifts the temporal perspective to a past reference time, which, as we saw above, may correlate with a distinct objective time. From this past perspective time, the expressed event is posterior; that is, it is a kind of future in the past. Correspondingly, conceptualized from this perspective time, the realization of the event is still possible. The present indicative is then the corresponding TAM choice. An important clue to substantiate our hypothesis is, as we will see, that the instances we find all pertain to oral or close-to-speech varieties (see Section 4.2) (see Wüest, 1993, p. 231 for the varying strength of correlation between temporal perspectivization and text types). We have already noted above that the phenomena at hand make use of secondary functions of the linguistic forms. In oral speech, exploiting the potential of flexibility of language is even more common (see also Labeau, 2006, p. 18-19).

## 2.4 The role of predictions and what we can learn from the data

When processing language, we partly resort to prestored knowledge in order to predict what is to come next (see Kuperberg and Jaeger, 2015 for a theoretical overview). Thus, prestored knowledge has an important function in communication. As Kuperberg (2013, p. 14) underlines, the "benefit of a predictive language processing architecture is comprehension efficiency". It may also be calculated, as shown, for instance, by Levy (2008), who focusses on surprisal (that is, basically, when predictions are not met). The research in the realm of predictions frequently discusses verbal properties and also touches upon the role of genre (see below). Importantly, as we will indicate below, we suggest an alternative way of approaching the phenomena, which takes genre to be a predictor of deviating TAM forms. Addressed in this way, football language is an especially revealing case. By analyzing corpus data containing the phenomena in question, we build a basis for future experimental research to profit from.

As discussed in the preceding sub-sections, the structures we analyze deviate from otherwise expected ones. This is especially interesting against the backdrop of predictive language processing. First, we might be inclined to ask how speakers would deal with deviating TAM forms if they interpreted them as errors. Hanulíková et al. (2012) indicate that when confronted with speakers with non-perfect acquisition of an L2, hearers do not show any reaction to syntactic errors (no P600 effect). Kuperberg (2013, p. 17) calls this "predictive error-based learning". One might be inclined to think that the same is happening in the case of the deviating TAM forms in football language. There are two main arguments against this view. First, as already indicated, the phenomena we are interested in are not extremely rare. Second, journalists and reporters are not completely free in their linguistic choices. We may assume that their employers would refuse to accept an overly individual or defective style of speaking or writing. By contrast, the use of a specific diastratical marking in order to indicate pertinence to a group (see Koch and Oesterreicher, 2011), such as the group of football fans, may be permissible or even desirable in order to reach a high number of listeners or readers. However, idiosyncratic markers should never be too strong, in the sense that the content of the utterance still needs to be comprehensible to the general public. This is another argument favoring licensing effects, as discussed in Section 2.3. Again, if a hearer / reader were to incur processing difficulties every time she / he is confronted with a non-standard TAM form, the form would quickly be banned from the genre it appears in. By contrast, as we find the phenomena in question frequently in football speech, they cannot be expected to be overly costly in terms of the comprehenders' processing. Rather, we assume a genre-based prediction that such TAM forms will occur.

What do we know about the role of TAM forms in predictive language? In the research literature, verbal properties are shown to play a crucial role in the predictive processing of language (see, for instance, Kuperberg and Jaeger, 2015, p. 6-7, and the references therein). Among the best studied verb-related factors are its argument properties, such as its selection restrictions (see Altmann and Kamide, 1999). However, tense and aspect have also been investigated (see Kuperberg and Jaeger, 2015, p. 7; Arai and Keller, 2013, p. 5; both with further references). For instance, Altmann and Kamide (2007) show that the tense-aspect morphology is relevant for predicting the verb's object. Furthermore, Philipp et al. (2017) discuss the interaction of semantic roles with telicity and event structure with respect to processing costs. Graf et al. (2017) analyze the interaction between verbal (telicity) and nominal (agentivity) features. As a final example, Dery and Koenig (2015) specifically focus on temporal updates. By contrast, mood and modality do not seem to have aroused much interest. However, genre, our second category, has also been studied (see Kuperberg and Jaeger, 2015, p. 16). For instance, according to Fine et al. (2013, p. 15), comprehenders are sensitive to genre and other more individual properties of the input, and they "continuously adapt their syntactic expectations" as they are exposed to new linguistic input. Further evidence can be found in Squires (2019). She discusses insights from several studies underlining that the knowledge of the listener about the speaker in terms of dialect or sociolect positively influences how possible expectation violations are processed (see Squires, 2019, p. 2-4, with further references). More specifically, she presents three experiments concerning the role the genre of pop songs has on the evaluation of non-standard morphosyntactic forms by listeners (see Squires, 2019, p. 8-23). She concludes "that speech genre can serve as an expectation-shifting sociolinguistic cue during sentence processing" (Squires, 2019, p. 23). However, according to Squires, the listeners' expectations are rather vague with respect to what phenomena might occur and how they deviate from standard language (see Squires, 2019, p. 23).

An important part of the research literature focusses on syntactic or morphological markers as predictors. By contrast, Fine et al. (2013) take syntactic structure as the predicted component. Similarly, we analyze TAM forms as predicted elements. To be precise, we argue that they are predicted on the grounds of the genre they occur in. The relevance of this idea is backed by Kuperberg and Jaeger (2015, p. 4, with reference to Anderson, 1990), who state that a comprehender will "use all her stored probabilistic knowledge, in combination with the preceding context, to process th[e] input". However, we do not analyze prediction locally, that is, in a sentence-based way, but globally in the sense of the classification of a whole discourse in terms of the subject it covers. We find a similar principle in the study by Schumacher and Avrutin (2011), who analyze the role of a certain discourse type for the processing of article-less noun phrases. The discourse type that they are interested in, a classification for which they use the category term "register" (see Schumacher and Avrutin, 2011, p. 306, footnote 3, for the way they determine it), is that of newspaper headlines. The authors present two studies involving NPs without an article, where the participants are only made aware of the pertinence of the critical items to a headline in one study, not in the other (Schumacher and Avrutin, 2011, p. 307). They show "that awareness of a particular register, and the expectations associated with it, has an impact on the readers' processing patterns" (Schumacher and Avrutin, 2011, p. 318). On these grounds, we assume that an analysis in our terms is promising.

## 3. Materials and Methods

As its definition is content-based, football language is a broad concept linguistically (see Burkhardt, 2006 for a distinction of three lexically motivated sub-types). Football-related content may be presented in very different situational (for instance, with regard to the aim of the speech or writing event), social (with respect to the speaker / hearer constellation) and medial (concerning the medium) contexts. Typical situations for football language are football news reports on the radio, live commentaries on television, newspaper articles or printed interviews, but also fans talking among themselves. Thus, the concept of football language needs to capture a possible diversification in terms of diaphasics, diastratics and diamedial realization (see Koch and Oesterreicher, 2011). We assume an oral predominance with regard to the general use of football language, which however, may be taken to be a general property of linguistic data (see Sinclair, 2005, section 5, who classifies the distribution as a general problem for corpora). However, due to practical issues we restrict ourselves to written data at this point, which may include direct quotes and spoken interviews published in a written format.

An important part of the insights generated comes from the data collection process and the challenges we encountered when retrieving the linguistic material from the corpora. Therefore, the present section goes into further details regarding the data analyzed. It is especially dedicated to the corpus studies we conducted, the factors we considered and how we solved the issues that arose.

In order to understand the phenomena in depth, we collected data from pertinent corpora. In both languages, the linguistic material was annotated upon retrieval. Although the phenomena analyzed, in their core, both boil down to deviant TAM forms, they do not pertain to directly parallel verbal paradigms. Due to the systemic differences, different factors need to be accounted for in their retrieval and in the ensuing analysis. On the one hand, in German, the syntactic structure is relevant, in the sense that the occurrences are combinations of a subordinate and a main clause. However, as the annotations of the corpora we used do not consider syntactic structure, we can only make use of it in the corpus queries in cases where an explicit subordination marker is realized or a specific word order is used (verb-first). By contrast, an implicit subordination is much more difficult to find. On the other hand, in French what is of most interest is basically the re-occurrence of a tense-aspect form in a sequence of sentences. But the information on the quantity of a form within the textual entity it occurs in (the sentence, the paragraph or a text in its entirety) cannot be directly retrieved from a standard online corpus.

An issue beyond the linguistic means of expression is the availability of specialized corpus data, which differed between the two languages. With respect to German, we had specialized corpora at our disposal (see Section 3.1). Interestingly however, relevant examples were not easy to find there, which might be considered a sign of low frequency. By contrast, we did not have specialized French corpora. Therefore, in order to collect the data, we used specific verbs and collocations (see Section 3.2). Furthermore, as we intended to analyze whole articles in French, we extended the passages retrieved step by step using the corresponding database or through queries on the general internet.

## 3.1 Collection of German data in specialized and non-specialized corpora

As already noted, the German phenomenon does not seem to have been described in the literature. Therefore, we started off with an unstructured study using the general internet (via [www.google.de](www.google.de)) in order to determine relevant factors for its occurrence. They concerned the syntactic structure and, more importantly, the relevant verbal types (see below). For the main study, we had corpora at our disposal which specifically contained German football language. Interestingly however, we only found very few instances of the phenomenon of interest. Therefore, we additionally used unspecialized corpus data to balance our findings. We analyzed sub-corpora contained in Cosmas II, a database issued by the Leibniz-Institut für Deutsche Sprache (IDS) ([https://www2.ids-mannheim.de/cosmas2/](https://www2.ids-mannheim.de/cosmas2/)). It comprises 570 sub-corpora with a total of 56.5 billion tokens ([https://www2.ids-mannheim.de/cosmas2/uebersicht.html](https://www2.ids-mannheim.de/cosmas2/uebersicht.html), accessed: June 16, 2021). The sub-corpora are organized into eighteen different archives ([https://www2.ids-mannheim.de/cosmas2/projekt/referenz/archive.html](https://www2.ids-mannheim.de/cosmas2/projekt/referenz/archive.html), accessed: June 16, 2021).

In a first step, we analyzed two sub-corpora of football language, which are both contained in the corpus "W - Archiv der geschriebenen Sprache". The first one is "KSP - Fußball-Spielberichte, kicker.de, 2006 - 2016". As the name indicates, it consists of a collection of data from the German football journal *Kicker*. Second, we investigated the sub-corpus "SID - Fußball-Liveticker, Sport-Informations-Dienst, 2010 - 2016". It consists of data from football life tickers. "KSP" amounts to 3,000 texts with 1.9 million tokens. "SID" contains approximately 1,800 texts with close to 3.8 million tokens. Even in the unstructured pre-study, we found a tendency for an oral predominance. However, there are two reasons why the two sub-corpora appear to be promising despite the fact that they present written texts. First, both may contain direct quotes. Second, the live ticker (SID) may be said to be rather close to the pole of a language of immediacy on the scale of linguistic conception by Koch and Oesterreicher (2012).

The corpus research with regard to the two sub-corpora had to be carried out in two steps. The first step exploited an important advantage of Cosmas II, namely that the entries are lemmatized and the corpus allows for relatively complex queries. As discussed in Section 2.1, the forms we are interested in are simple present tense forms (or share the form of the present indicative). As a consequence, we had to deal with a considerable amount of noise. Therefore, in the second step, we went through the hits manually and retrieved the relevant examples in order to analyze them more in depth.

In formulating the corpus queries we took into consideration the two different syntactic possibilities, namely, (i) clauses introduced by an explicit conjunction *wenn* ('if'), or (ii) clauses displaying verb-first word order. For the case of an overt conjunction, we used the query that the verb should follow within a range of five words from the conjunction. We intended the verbal inflection not to be restricted. Thus, we used the query, "wenn /+w5 &verb", where we filled the verb slot with a specific lexical item (see below). The second query type specified a verb-first clause (see Section 2.1). In Cosmas II, this can be spelled out as "&verb /w0 <sa>", which determines that the verb should occur as the beginning of a new sentence.

In the previously conducted unstructured analysis using the search engine google ([www.google.de](www.google.de)), we had intended to find out what verbs may occur in such contexts. There, we used simple co-occurrence patterns of different verbs in the present tense and denotations of individuals typically involved in football activities like *Torwart* ('goalkeeper') and *Schiedsrichter* ('referee'). The findings indicated that at least two factors were relevant. First, we only found verbs expressing typical actions in football matches, especially if they were compatible with rather colloquial football language. Due to its seeming frequency, an appropriate example is *machen* ('make') combined with the short demonstrative pronoun (*den*) in the structure *den machen* ('to score a goal (as part of a specific opportunity)'). Second, all the verbs we found were telic.

On these grounds we chose nine different verbs for our structured corpus study, among them *flanken* ('to center'), *halten* ('to stop (a ball); to save a penalty') and *verwandeln* ('to convert (for instance, a penalty)'). All of these verbs are typical of football contexts. In the uses we looked for, they are all telic or ingressive and most have a punctual reading. We considered all of them in both syntactic configurations. This led to a total of five relevant hits. The amount of noise was considerable. In "KSP", there were 423 hits, of which, however, none was relevant. In "SID", there were 380 hits containing the five relevant ones.

In a second step, we chose the sub-corpus of a local newspaper with a relatively high circulation, namely the *Kölner Stadt-Anzeiger*. This sub-corpus pertains to a group of newspapers which is often listed among Germany's top ten regional newspapers (see, for instance, for the second quarter of 2019 [https://meedia.de/2019/07/22/die-auflagen-bilanz-der-groessten-83-regionalzeitungen-kaum-noch-titel-mit-einem-minus-unter-2/](https://meedia.de/2019/07/22/die-auflagen-bilanz-der-groessten-83-regionalzeitungen-kaum-noch-titel-mit-einem-minus-unter-2/), accessed: June 16, 2021). The sub-corpus "KSA - Kölner-Stadtanzeiger, 2000 - 2019" is contained in "W2 - Archiv der geschriebenen Sprache". As its name indicates, it contains twenty years of the newspaper, amounting to over two million texts with more than six hundred million tokens. Again, we used both query types presented above, but we reduced the target verbs to four. We chose verbs which had a high probability, in terms of their lexical content, of occurring in such contexts. They express scoring (*machen*, 'to make', see above, *treffen*, 'to hit', *verwandeln*, 'to convert') or saving a goal (*halten*, 'to save') and are all compatible with a colloquial style. We restricted the study to the first one hundred occurrences per query. The resulting eight hundred hits paralleled the amount of data from the specialized corpora. In the same vein, the hits were examined manually in order to retrieve the relevant cases. Again, there was a high proportion of noise. Thus, although at least some of the hits presented football content, only three of the eight hundred hits are relevant in our terms. Interestingly, all three instances present direct speech. This matched our expectation based on the preliminary unstructured studies. By contrast, the hits from the life tickers (SID) were not instances of direct speech. However, they pertain to a close-to-speech variety (see Section 4.2).

## 3.2 Collection of French data in non-specialized corpora

As indicated in Section 3.1, for French we did not have specific corpora of football language at our disposal. Therefore, we slightly adapted our procedure. We maintained lexical meaning as a central component and also used verbs expressing typical football actions.

Furthermore, we made use of collocations (see Lehecka, 2015). We retrieved the relevant hits and their context from the corpora. Furthermore, we augmented the context as much as possible by using queries with strings from the preceding or the following context, either within the database or on the general internet.

With respect to French, we investigated newspaper data from the two data bases Emolex and Sketch Engine. We collected relatively long strings of context as we intended to find out more about the discursive functioning of the narrative imperfect. More specifically, our aim was three-fold. First, we collected sequential data in order to analyze the discourse context. In this respect, we investigated, for instance, the positioning of the form in question within a paragraph and what other tense-aspect forms it co-occurs with. Second, we analyzed the interaction of different imperfect uses in co-occurring contexts. Third, we were interested in the pervasiveness of the form in the text type at hand. While some of the insights from the first two aims are presented in Egetenmeyer (in press), the last issue is most relevant for the present paper.

We began with an analysis of the newspaper data contained in Emolex (see Diwersy et al., 2014) (the original URL, http://emolex.u-grenoble3.fr, is no longer available and has been changed to http://phraseotext.univ-grenoble-alpes.fr/emoBase/, accessed: June 16, 2021). It is a monolingual press corpus containing data from national and regional newspapers from the years 2007-2008. Kern and Grutschus (2014, p. 188) illustrate the two groups with *Le Monde* from 2008 and *Libération* from 2007, contrasted with *Ouest-France* from 2007 and 2008. In total, the corpus contains over 112 million words from close to 300,000 texts (see http://phraseotext.univ-grenoble-alpes.fr/emoConc/emoConc.new.php). Its data coverage is thus especially good. An important advantage of the corpus is that it contains the whole articles and does not inhibit their retrieval piece by piece in their entirety.

We conducted two different studies. The first one concerned nine verbal types or verbal phrases, where the verbs involved were marked for third person singular, *imparfait d'indicatif*. In terms of lexical context, the verbs were likely to occur in a football context. Among them were *centrer* ('to center'), *dévier* ('to deflect') and the collocation *marquer + but* ('to score + goal'). As *centrer* ('to center') presented a considerably higher number of instances than the other verbs used in the queries, we restricted our analysis in this case to the first 50 occurrences with football content. We found relevant instances for five of the verbs and retrieved forty-four relevant examples from a total of forty-two different newspaper articles. As indicated above, we were able to retrieve the whole texts in all cases.

Interestingly, the verb choice in combination with the inflection yielded a low percentage of noise in this corpus study. Approximately 83% of all hits occurred in a football context. Among these hits, the verbs contained in our queries may be classified in nearly 42% of cases as narrative uses of the imperfect in the sense defined in Egetenmeyer (in press) for football language. That is, they were part of strings of verbs marked by the *imparfait* which expressed sequences of events. Other instances showed a descriptive or background reading and the like, and were therefore not included in our data collection.

In order to test whether we could replicate our successful first study with another database, we issued further queries in the database Sketch Engine, and more specifically, within the sub-corpus "Timestamped JSI web corpus 2014-2017 French". The corpus is very large. Therefore, we restricted the data to the year 2016, leaving us with over one billion tokens. We looked for the two verbs *contrer* ('to counter') and *dribbler* ('to dribble' / 'to pass someone dribbling') and the verbal collocation *marquer + but* ('to score + goal'), whose components had to occur within a range of five words. Due to the high amounts of hits in all queries, we limited our focus to the first fifty items with a football context and where the noun *but* occurred in the singular. Thereby, we retrieved thirty relevant instances from twenty-nine different texts.

In twenty-two cases we were also able to retrieve the entire article, either from the database or from the free internet, using queries in Google. Again, we analyzed the data retrieved in more depth.

# 4. Results

In the previous section we discussed the ways in which we collected data on the linguistic phenomena presented in Section 2. These phenomena are formally different, although they are based on a similar conceptual principle. Due to the different languages we investigated, we also used different databases and corpora. The databases differ with respect to content-related specificity and with regard to the corpus query language, that is, what kinds of queries are possible. Therefore, we adapted our queries and we were able to retrieve relevant data in all cases. We then analyzed the data retrieved in depth.

It is important to recall the paramount aim of the contribution. We intend to collect evidence for the role genre plays in building up predictions for TAM marking. To achieve this goal, we need to understand the phenomena better. This guides the presentation of the results. In the following two sub-sections, we make reference to both German and French. Furthermore, both sub-sections combine quantitative and qualitative insights. Section 4.1 considers Aktionsart properties of the verbs involved in the deviant TAM marking. Section 4.2 discusses how the structures are embedded into the discourse they occur in.

## 4.1 Aktionsart properties

Our interest concerns morphological marking on the verb, tense-aspect and mood marking, which deviate from a basic prescriptive marking. As indicated in Section 2.3, the phenomena in both languages show an important parallel in their temporal anchoring, resulting from the temporal perspective involved. Therefore, we can determine the most direct and pervasive potential factor of the forms analyzed as being the part of the verbs' lexical meaning which is relevant in terms of temporal structure, namely, the Aktionsart of the verbs.

As part of the queries, we controlled for the Aktionsart properties. In the German queries, all verbs we searched for were telic in the reading in question. The examples we retrieved consist of seven achievements and one accomplishment. In addition, the structures contained a second verb, where we found a certain variance in terms of Aktionsart. The data retrieved show two achievements, two accomplishments, one activity and three states. This variance is also relevant for the relationships holding within the structures (see Section 4.2).

Although in our French queries one verb was atelic, we were only able to retrieve instances with telic verbs. In the most extensive study we realized with Emolex, we were able to retrieve examples containing five different verbs of which, in their typical reading, three express achievements, one an accomplishment and one an ingressive process. Of the achievement verbs, *centrer* ('to center') yielded most relevant examples (19), followed by *dévier* ('to deflect') (9) and *marquer + but* ('to score + goal') (8). The queries with the accomplishment verb *dribbler* ('to dribble' in the sense of 'to pass someone by dribbling') and the ingressive process *contrer* ('to counter') both yielded four relevant occurrences each.

As noted in Section 3.2, in this study, we retrieved whole articles and analyzed them further. The forty-four hits pertained to forty-two articles. From these forty-two articles, we selected those where at least 75% of all inflected verbs, not counting direct discourse if it occurred, were marked by the *imparfait*. In these cases, the *imparfait* forms typically occur in long sequences in which other forms intervene only very rarely. The non-imperfective forms rather tend to occur at the beginning or at the end of the articles. This augments the probability of narrative uses in sequence. The data set contained two articles which did not involve any other finite form than the *imparfait* and a further twenty-three articles showed 75% or more *imparfait* markings. Four further articles were close to the threshold, but we excluded them from the further step, together with the articles with fewer *imparfait* verbs than this. As part of this step, we analyzed the Aktionsart of all the *imparfait* verbs contained in the subset of twenty-five articles. In cases of doubt on the classification of the Aktionsart, we took a bearing on Lehmann (1991). We thereby intended to test the finding of Bres (1999, p. 5) that narrative imperfects frequently occur with lexically bounded verbs. However, it is important to note that not all of these *imparfait* verbs show a narrative use. Although a very high share actually occurs in chains of *imparfait* verbs, some may express intervening descriptions, fulfilling the function of stage setting or other non-narrative functions. Habituals are also possible. However, as the sheer proportion of verbs marked by the *imparfait* is unprecedented in French texts, it is reasonable to maintain this broad focus and not to select specific uses at this point.

In the twenty-five texts identified, we found a total of 599 inflected verbs. 499 of these verbs are marked by the *imparfait*. Of these verbs, 316 are telic (63.3%) and 183 are atelic (36.7%). By far the largest group is that of achievement verbs, with 227 (45.5% of all verbs marked by the *imparfait*). This is precisely what we may expect from a football report, namely that events in sequence are narrated. The group with the second largest share is that of states, with 128 verbs (25.7%). Again, given the way we counted the verbs, this may be expected. Many of these verbs contribute background information against which the importance of the narrated events is to be evaluated. Furthermore, there are 69 accomplishment verbs, 20 verbs expressing an ingressive process, which we also counted as telic, and 55 activity verbs.

As noted with respect to the data coming from Sketch Engine, we were not able to retrieve the entire article in all cases. Therefore, we did not repeat the extensive study of the Aktionsart of all *imparfait* verbs in the articles, as we did with the data retrieved from Emolex. However, we analyzed the data with respect to preceding, following and intervening non-*imparfait* tense-aspect forms (see Section 4.2).

## 4.2 Discursive properties

In the preceding sub-section, we have already presented insights from the French articles which we retrieved as a whole. This was a first glance with respect to the more global perspective. However, in what follows, we present insights which concern the relational structure of the forms within their co-text. There are four main discursive properties which we derived from the data. With respect to German, we analyzed the function that the structure plays in the discourse context and the rhetorical relation which holds between its components. With regard to French, we investigated what tense-aspect forms occurred before and after long strings of verbs marked by the *imparfait*. Furthermore, we investigated whether another form intervened and if so, which form that was. These insights may also be understood as indications of the discursive anchoring of the structures in question.

We expected the German present indicative substituting a pluperfect subjunctive to occur in oral or close-to-speech varieties. Interestingly, none of the cases we found in the corpora with specific football language are direct discourse (five instances), while in the non-specific newspaper corpus all three hits occur as part of direct speech. As noted with respect to the specialized corpora, we retrieved all five instances from the corpus presenting live ticker data ("SID"), which is a variety with characteristics of a language of proximity. Thus, the distribution of the data is as we expected.

Furthermore, we analyzed the rhetorical relation between the two clauses comprising the German structure (see Asher and Lascarides, 2003; Kehler, 2002 and others). Interestingly, relationships involving temporal sequentiality predominate. Four instances show a strict contiguity between the two eventualities expressed, which may be best classified as cases of OCCASION in terms of Kehler (2011, p. 1970) (see example 5). And, if we were to make such a fine-grained distinction, three further instances express a less direct temporal sequence that may be classified as NARRATION in the sense of Asher and Lascarides (2003, p. 162) (see example 6). One example shows an ELABORATION relation (see example 7).

(5) *Der Peruaner kommt an den Fünfer gerauscht, verpasst den Ball aber ganz knapp [...]. Wenn er den _trifft_, dann _zappelt_ das Leder auch im Netz.* (Cosmas II: SID/B16.00096)

'The Peruvian rushes to the goal area, but just misses the ball. If he had hit (lit.: hits) it, the leather would have wriggled (lit.: wriggles) in the net.'

(6) *Nach 70 Minuten hätte Nhu-Phan Nguyen den VfL erneut in Führung bringen können, scheiterte jedoch [...] an FCB-Torhüter Kevin Kraus. „Womöglich war das der Knackpunkt. _Verwandelt_ er, dann _gewinnen_ wir das Spiel, da bin ich mir sicher", erklärte Brunetto.* (Cosmas II: KSA14/SEP.08133)

'After 70 minutes, Nhu-Phan Nguyen could have given VfL the lead again, but failed to beat FCB goalkeeper Kevin Kraus. "Possibly that was the crux. If he had converted (lit.: converts), then we would have won (lit.: win) the game, I'm sure," explained Brunetto.'

(7) [1] *Schäfer fausted einen Groß-Freistoß genau vor die Füße von Lex,* [2] *der aus etwa 16 Metern Maß nimmt* [3] *und mit Gewalt drauf hält.* [4] *Geht er rein,* [5] *schießt er das Tor des Monats –* [6] *doch so geht die Kugel über den Querbalken.* (Cosmas II: SID/Z15.00097)

'[1] With his fist, Schäfer diverts a free kick by Groß right at the feet of Lex, [2] who takes aim from about 16 meters [3] and shoots forcefully. [4] If the ball had gone (lit.: goes) in, [5] he had scored (lit.: scores) the goal of the month – [6] but this way the ball flies over the crossbar.'

Example (7) also underlines what we argued for in Section 2.3, namely, the conceptual proximity between the true futurate present tense and the present indicative substituting a pluperfect subjunctive. Here, the live ticker seems to present the information as if it were in objective real-time. At the time of the shot, the goal is still possible. Thus, uttered strictly at this time, a present tense form may express potentialis or future reference. However, even if we abstract away from the fact that an utterance (in written form) is impossible in objective real-time (perception and taking notes simply takes too much time) and take the relevant temporal measurement to be a subjective time flow, we may still classify the second sentence of the example ([4], [5]) as irrealis. This is so because of the evaluation in clause [5] (*schießt er das Tor des Monats*, 'he scores the goal of the month'), which can only be ascribed after the realization of the shot. And at that point in time, it is already known to the speaker that the attempt has been in vain.

In the French data, we analyzed the tense-aspect forms occurring before the sequences of verbs marked by the *imparfait*. In this part of the study, we included all examples retrieved. We analyzed both the data from Emolex and the data from Sketch Engine. The sequences of verbs, typically one or more paragraphs, consisted of or at least contained the narrative uses of the *imparfait* which had been the focus of the data collection.

As noted, in the data retrieved from Emolex, in two cases two relevant verbs are contained in the same article. As, in both of these cases, the two verbs also pertain to the same sequence of *imparfait* verbs, we do not count two separate instances (which would amount to four); this leaves us with 42 relevant texts. 12 cases actually show an *imparfait* as first verb. Most frequently, the preceding verb shows a past tense with indicative mood (23 cases, 54.8% of all cases and 76.7% of the cases showing a verb other than an *imparfait* before the sequence). This may be expected. However, within this group the distribution does not adhere to a clear principle. There are eight verbs marked by the *plus-que-parfait* (the pluperfect), six marked by the *passé simple* (simple past marked for perfectivity) and nine marked by the *passé composé* (compound past). The remaining preceding verbs comprise four verbs marked by the present tense, one by the present conditional and two by the past conditional. With respect to the tense-aspect forms following the sequence of verbs marked by the *imparfait*, there is no clear tendency. The largest share is composed of the cases where no verb with a different tense-aspect form follows (19 cases). Apart from this group, there are three verbs marked by the pluperfect, four marked by the *passé simple* and another four by the *passé composé*. Nine verbs are marked by the present tense, two by the past conditional and one by the simple future. Finally, we analyzed the persistence of the *imparfait* within the largest sequence of *imparfait* forms in the articles. In our data from the Emolex corpus, the *imparfait* was predominantly persistent, as 26 "chains" of *imparfait* verbs were not interrupted at all (61.9%). In the 16 cases with an interruption (38.1%), we analyzed the verb forms intervening in the chains. More specifically, we focused on the first interrupting verb form. The variance is conspicuously reduced. There are only five different TAM forms, of which only three occur more often than once. Five verbs are marked by the pluperfect, four by the present tense, five by the past conditional and one each by the *futur antérieur* (future perfect) and the present subjunctive. In this data set, the interruption of the chain of *imparfait* verbs is mainly realized by only one verb. By contrast, there are only three cases where the interruption comprises more than one verb, and in one further case, the chain of *imparfait* verbs is interrupted twice. As stated, we only counted the first interrupting tense-aspect form. Furthermore, it is interesting to note that many of the intervening tense-aspect forms are typical indicators of a perspective shift, namely, the pluperfect and the present tense (see Becker et al., 2021). This is another indication favoring our hypothesis of a special perspective in the case of the footballer's *imparfait* (see Section 2.3), as interruptions of the chains correlate with a shift in perspective (see also Sthioul, 2000 for the interplay of tense-aspect forms and perspective taking).

The linguistic material which we retrieved from Sketch Engine of 29 texts partly confirms our findings from the above-described set of data. As noted, we were only able to collect the entire articles in 22 cases, which needs to be taken into consideration in the analysis and the results. We repeated our analysis of tense-aspect forms preceding and following the main sequences of *imparfait* verbs ("chains"). With respect to the preceding tense-aspect forms, the tendency found in Emolex of a predominance of past tense-aspect forms did not repeat itself. Six articles start with the *imparfait* immediately. In one of these cases, however, the preceding paragraph is missing. Of the remaining 23 instances, less than half show a past tense (11 cases, 47.8%), so that non-past tenses are slightly dominant (12 instances, 52.2%). The first sub-set consists of two verbs marked by the pluperfect, two by the *passé simple* and seven by the compound past. In the second sub-set, the present tense dominates with 11 forms. The remaining verb is marked by the simple future. Similarly to the data from Emolex, the data from Sketch Engine also lacks a clear tendency in the realm of the tense-aspect form following the *imparfait* chain. However, the variance is reduced. In this case, seven instances lack a following verb not marked by the *imparfait*. However, in three of these cases a following paragraph is missing. Of the remaining 22 verbs, 12 are marked for past tense (54.5%) and ten show a non-past tense-aspect marking (45.5%). More specifically, two verbs are marked by the pluperfect, one is marked by the *passé simple* and nine are marked for the compound past. Again, the present tense is predominant in the non-past array as nine verbs show a present indicative marking and only one is marked by the simple future. Relatively speaking, the *imparfait* chain is interrupted more frequently in this data set, namely in 14 cases (48.28%), as opposed to 15 examples without interruption (51.72%). Interestingly enough, among the interruption cases, the two tense-aspect forms with high perspective shifting potential, the pluperfect and the present tense, are again predominant. The pluperfect occurs five times and the present tense seven times. Furthermore, there is one instance of a compound past and one case of simple future. However, it is important to point out a restriction on these last results. In ten articles, the *imparfait* chain is interrupted more than once. We analyzed only the first interrupting verb, which, in contrast to the first data set, always consists of only one non-*imparfait* verb.

## 5. Discussion

In the paper at hand, we reported on several corpus studies of football language concerning TAM forms which differ from the expected standard forms. In German football talk, the present indicative may replace the pluperfect subjunctive. In French football reports, the imperfective past may replace the perfective past. On the basis of our results, we are able to address the basic research questions we listed in the Introduction. (i) With regard to frequency and distribution, we collected evidence for the established status of the uses as they occurred in different corpora. The French TAM form appears to be more frequent than the German counterpart. The French footballers' *imparfait* can be found across longer strings of text. In addition, the German TAM deviation has been shown to occur in oral and close-to-speech varieties, while the French one pertains to written language. (ii) The grammaticalization status of these phenomena was underlined in terms of their lexical variability and also with regard to their discourse semantic variability. For both languages, we showed that a wide range of Aktionsart features is compatible with the phenomena, although we used only telic verbs in the German corpus queries. When analyzing the whole newspaper articles (or the retrievable part), French also displayed a predominance of telic verbs. This is in accordance with Bres (1999, p. 5) and others who analyze the literary use of the narrative imperfect (see also below). Furthermore, for German, we analyzed the rhetorical relations within the structure, which also show a certain variability. The detailed analyses were necessary as the previous research literature is still poor in at least the following two regards. First, it does not seem to have accounted for the German present indicative replacing the pluperfect subjunctive in football talk at all. Second, with respect to French football language, the literature is focussed on relatively short excerpts or discusses the distribution of tense-aspect forms in football reports in general (see for instance work by Labeau, 2004, 2007). By contrast, we included longer strings and also addressed entire football articles as a whole. Furthermore, the combined analysis of the two phenomena is new.

The other research questions go beyond the quantitative analysis of the data. (iii) Why do speakers use such deviant TAM forms and, relatedly, (iv) what linguistic and non-linguistic clues do we find which serve as coping mechanisms on the part of the comprehender? As we saw in Section 2.3, there are good arguments that the forms analyzed function on the grounds of a perspective shift. We can only hypothesize about the reasons why speakers do so. However, given the content referred to, we may assume that the speaker has the intention to induce interest in the listener. Therefore, it is probable that the speaker intends to convey a conceptual proximity with respect to the content (see Walker, 2008, p. 300 for arguments against a so-called hot news reading of the present perfect in English football language). Importantly, as we saw in Section 2.3, this "flexibility" in the use of TAM forms seems to be licensed by the linguistic and extra-linguistic properties of the genre. More specifically, the most central subject, football matches, adheres to a specific script or frame with typical events following a well-entrenched course of action and which are in accordance with a set of rules. Furthermore, the events may be determined temporally in terms of objective time. The speaker may assume that the comprehender has all this knowledge. Therefore, the temporal (French) and world reference indices (German) may easily be retrieved contextually. As a consequence, the speaker may resort to secondary functions of the TAM forms. Apart from scripts and frames and objective time, the comprehender may rely on linguistic means to decipher the message. This is especially relevant in French, where the lack of perfectivity is, at least partly, compensated for by lexical boundedness (see also Bres, 1999, p. 5). We saw that the deviation in French may occur across whole newspaper articles. By contrast, due to its properties, the German phenomenon seems to be restricted mainly to referring to single events in the extra-linguistic world. It should be noted that our data indicates a very specific quality of these events. The events often seem to be decisive for the match in question. Thus, they are prominent within the discourse situation.

Finally, the paramount research question was (v) what insights can we gain from the phenomena in predictive language processing? We described the deviant TAM forms in two languages as a function of genre. As we have seen, the specific properties of the genre license the deviant uses in the first place and also convey the necessary parameters to decipher the intended message. Schumacher and Avrutin (2011) present a similar approach, although the phenomenon they discuss pertains to the realm of reference to individuals and not to events as in our case. Thus, the case of deviant TAM forms in German and French football language shows the extensive predictive potential of genre.

To conclude, it should be mentioned again that our studies were concerned with corpus data. They allowed us to explore the whole range of properties of the phenomena discussed. However, in a further step, it should be investigated whether our insights and our hypotheses concerning processing bear closer examination. Therefore, they should be tested with experimental methods. Apart from a more general analysis, following, for instance, the example by Schumacher and Avrutin (2011) or Squires (2019), in order to confirm the role of genre in the processing of the deviant TAM forms, it would be especially interesting to test the role of the linguistic properties we determined. This might concern the preference for telic Aktionsart categories in French or, for example, the range of rhetorical relations in the German structures. The findings would also be of relevance for the functioning of these categories beyond the language of football. Furthermore, the description of the German present indicative replacing the pluperfect subjunctive should be broadened. Of special interest is the grammaticalization path. It could be tested whether genericity might have functioned as a gateway, as this was a factor that led to some confusion for our assistants. Finally, the analysis of the German TAM use in football language should be extended to other structures. As it seems, there are further interesting deviations to be detected.

## Funding

The research was funded by the Deutsche Forschungsgemeinschaft (DFG, German Research Foundation) as part of the Collaborative Research Center SFB 1252 *Prominence in Language*, Project-ID 281511265, which is gratefully acknowledged.

## Acknowledgments

I thank Mary Chambers for proof-reading, Barbara Zeyer, Laila López Armbruster and Vanessa Reibaldi for practical assistance and Franziska Kretzschmar for an open ear and input concerning an experimental view on the problem which, however, I have not pursued at this point. I thank Olivier Kraif for explaining to me how Emolex can now be accessed. I thank the audience at a talk in Neuchâtel in June 2018 at the conference Chronos 13 for their feedback concerning football language. I thank Martin Becker, Mark Ellison, Klaus von Heusinger, and Petra Schumacher for their input following a talk on some of the data from this paper given in Cologne in June 2021. Finally, I thank my football-interested friends for discussing with me the "strange verbal inflections" used by German commentators on live football matches on national television.

## Corpus

### Online data bases

### Further data cited in this contribution